\documentclass[letterpaper, 10 pt, conference]{ieeeconf}  
\IEEEoverridecommandlockouts
\usepackage{cite}
\usepackage{amsmath,amssymb,amsfonts}
\usepackage[ruled,vlined,linesnumbered,noend]{algorithm2e}
\usepackage{graphicx}
\usepackage{textcomp}
\usepackage{xcolor}
\usepackage{makecell}
\usepackage[font=footnotesize]{caption}
\usepackage{subcaption}
\newcommand{\mbf}{\mathbf}
\newcommand{\bs}{\boldsymbol}

\title{\LARGE \bf TrajectoTree: Trajectory Optimization Meets Tree Search for Planning Multi-contact Dexterous Manipulation}

\begin{document}
\author{ Claire Chen$^{1}$, Preston Culbertson$^{2}$, Marion Lepert$^{2}$, Mac Schwager$^{3}$, and Jeannette Bohg$^{1}$ 
\thanks{This research was supported in part by NSF NRI grant IIS-1925030.}
\thanks{
The authors are with Stanford University, Departments of $^{1}$Computer Science, $^{2}$Mechanical Engineering, and $^{3}$Aeronautics and Astronautics. \{clairech, pculbertson, lepertm, schwager, bohg\}@stanford.edu.}
}

\maketitle
\thispagestyle{empty}
\pagestyle{empty}

\begin{abstract}
Dexterous manipulation tasks often require contact switching, where fingers make and break contact with the object. We propose a method that plans trajectories for dexterous manipulation tasks involving contact switching using contact-implicit trajectory optimization (CITO) augmented with a high-level discrete contact sequence planner. We first use the high-level planner to find a sequence of finger contact switches given a desired object trajectory. With this contact sequence plan, we impose additional constraints in the CITO problem. We show that our method finds trajectories approximately 7 times faster than a general CITO baseline for a four-finger planar manipulation scenario. Furthermore, when executing the planned trajectories in a full dynamics simulator, we are able to more closely track the object pose trajectories planned by our method than those planned by the baselines. 
\end{abstract}

\section{Introduction}
Dexterous manipulation tasks, which consist of multiple fingers cooperating to manipulate an object \cite{Okamura2000}, often require fingers to make and break contact with the object.
For example, a multi-fingered hand manipulating a cube with a fixed grasp will only be able to rotate the cube by so much before the fingers reach their workspace limits; to rotate the cube any farther requires the fingers to switch contacts. Planning such tasks requires finding not only a discrete sequence of contact switches, but also the piecewise continuous trajectories that satisfy dynamic equations and force constraints for each part of the sequence.

Contact-implicit trajectory optimization (CITO) methods jointly find trajectories for state, input, and contact forces, making them a good candidate for planning finger joint configuration, torque, and contact force trajectories for dexterous manipulation tasks. The most straightforward way to apply CITO to such tasks is to use a general formulation, such as the one presented in \cite{PosaCIO}, with the appropriate dynamics constraints. We consider a planar task of manipulating a rectangular object in the gravity plane with four fingers, each with two degrees of freedom. When using a general CITO formulation to plan this task, we find that it not only requires long solve times, but also often plans trajectories that are highly dynamic, resulting in more frequently dropping the object when being executed with a low-level tracking controller. One could obtain more robust dexterous manipulation trajectories with CITO by introducing additional constraints into the optimization problem, such as requiring the object to always be stably grasped. While the CITO framework is amenable to such constraints, enforcing, for example, a grasp stability constraint would require selecting a combination of fingers to remain on the object at each time. Consequently, the optimizer faces a fundamentally discrete choice at each time, which is difficult to optimize whether modeled using continuous constraints or integer variables. 

\begin{figure}[t]
    \centering
    \includegraphics[width=1\linewidth]{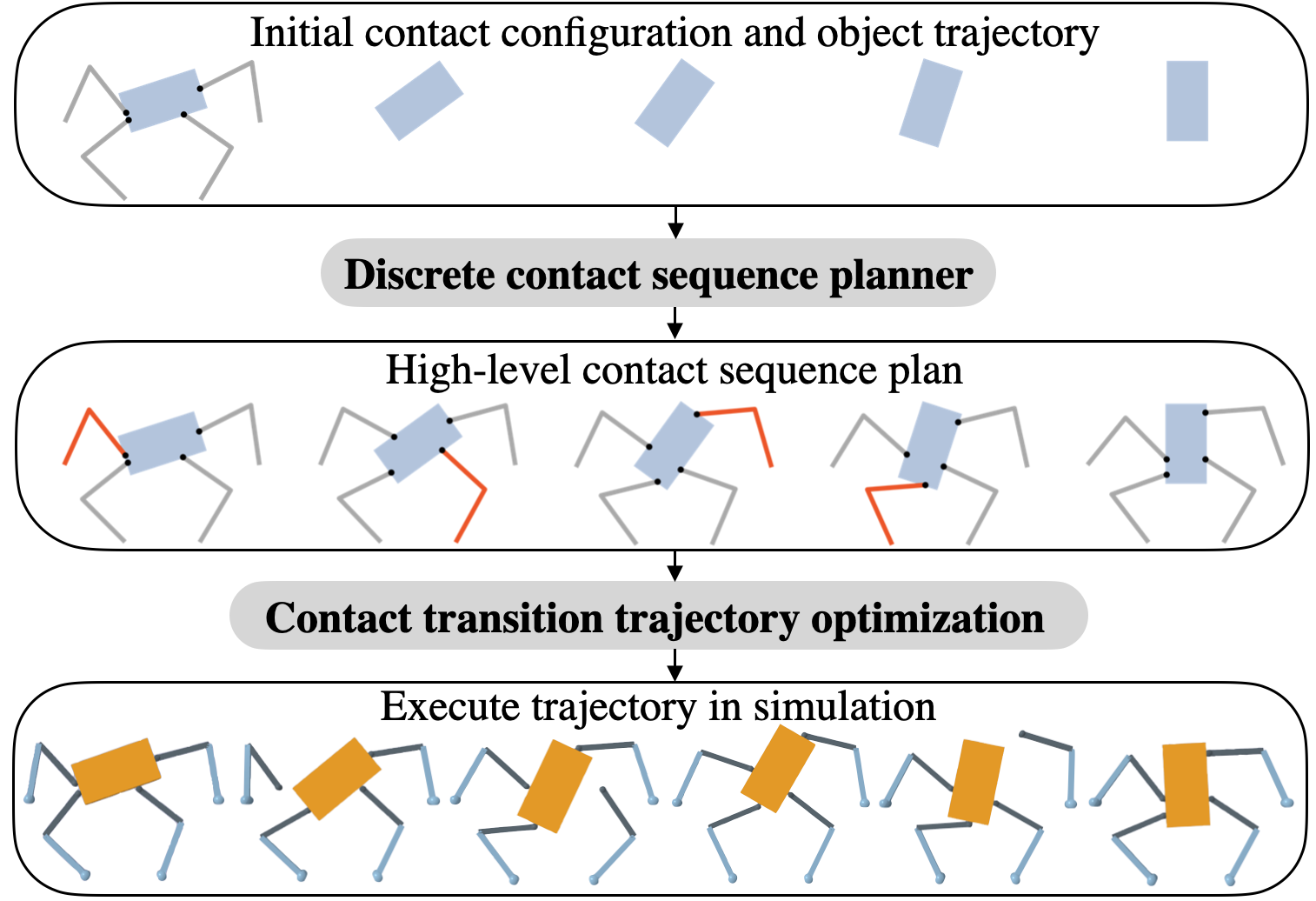}
    \caption{An overview of TrajectoTree for planning an object rotation task with four contact switches. Given an initial contact configuration and desired object trajectory, TrajectoTree use a discrete planner to find a feasible contact sequence, with fingers that switch contacts drawn in red. It then uses CITO with constraints from the high-level plan to plan the full trajectory.}
    \label{fig:method}
\end{figure}

Instead, given a desired object trajectory, we propose to first find a sequence of contact switches for the fingers using a discrete planner that reasons about kinematics and grasp stability. We then use this contact sequence to either improve the initial guess or impose additional constraints in the CITO. Specifically, we constrain the contact point locations to those planned by the contact sequence planner and only allow one finger to make and break contact with the object at any given time. Although we fix some contact point locations, we still model the contact dynamics of free fingers with complementarity constraints, allowing them to make contact with the object at any time during a contact transition.

We call our algorithm, which combines high-level tree search with low-level CITO, TrajectoTree. We show that TrajectoTree not only plans dexterous manipulation trajectories robust against dropping the object using CITO, but also improves solve times. Specifically, for a four-finger planar manipulation scenario, TrajectoTree has planning times that are, on average, approximately 7 times faster than a general CITO baseline. Furthermore, when executing the planned trajectories in a PyBullet simulation environment, a low-level finger tracking controller is able to more closely execute the object trajectories planned by TrajectoTree than trajectories planned by other CITO baselines.

\section{Related Work}
This work uses CITO as a strategy for dexterous manipulation planning. While both CITO and dexterous manipulation have been studied extensively, contact-implicit approaches to manipulation are relatively rare.
As such, we provide a fairly disjoint discussion of these two bodies of work, highlighting the few existing works which combine them. Given that dexterous manipulation relates to this work at the task level, we provide a general summary of some commonly used approaches. In contrast, this work focuses specifically on contact planning, warranting a more detailed analysis of individual prior contact planning methods from both the manipulation and locomotion communities.

\subsection{Contact-implicit trajectory optimization}

Trajectory optimization is a key method for planning robot motion, generating motion plans that locally optimize a given cost functional, subject to a set of constraints. Although trajectory optimization generalizes well to different systems, traditional methods can only handle smooth dynamics, and not the discontinuous dynamics introduced by contact switching. In recent years, several works have introduced a class of trajectory optimization methods that address this limitation. These methods, termed CITO, simultaneously plan state, control input, and contact force trajectories without needing a pre-specified contact mode schedule. They handle the hybrid dynamics of contact with either complementarity constraints \cite{Stewart2000, Yunt2006,PosaCIO,Manchester2017,Landry2019BilevelOF} or with soft constraints implemented as a penalty term in the cost function \cite{Mordatch2012DiscoveryOC, Mordatch2012ContactinvariantOF}. In this work, we follow the formulation introduced in \cite{PosaCIO}. 

Since the introduction of CITO methods, many works have applied them to whole-body dynamic motion planning \cite{Dai2014, Marcucci2017ATT}, quadruped locomotion \cite{Landry2019BilevelOF}, and single-leg jumping \cite{Mastalli2016HierarchicalPO}. 
Of these works, \cite{Marcucci2017ATT} and \cite{Mastalli2016HierarchicalPO} use hierarchical planning schemes, first planning a trajectory with a simplified robot model and then using this to warm-start the full trajectory optimization. TrajectoTree adopts a similar hierarchical design strategy, first considering a simpler kinematics problem before introducing full dynamics.
More recently, CITO methods have also been used for planning non-prehensile object manipulation \cite{Sleiman2019, Onol2019ContactImplicitTO}. There have also been a few works that use CITO for planning dexterous manipulation tasks \cite{Mordatch2012ContactinvariantOF, Gabiccini2018CIODexMan}. 
In \cite{Mordatch2012ContactinvariantOF}, the authors impose both dynamics and contact constraints using a soft penalty formulation to synthesize manipulation motions for computer graphics. The method proposed in \cite{Gabiccini2018CIODexMan} is more similar to ours, in that it uses a complementarity constraint formulation. While both methods are able to plan dexterous manipulation tasks that involve contact switching, results are only shown as animated visualizations of the trajectories, and do not include discussions of the planning times.

While one might argue that the versatility of CITO makes it trivial to apply to planning dexterous manipulation tasks, in reality, obtaining good solutions with reasonable solve times still depends on providing an appropriate initialization and task-specific constraints. TrajectoTree accomplishes this by augmenting CITO with an additional high-level planner.

\subsection{Planning dexterous manipulation}

Contact switching is an integral part of dexterous manipulation, but reasoning about how to make and break contact with an object is difficult due to the combinatorics and hybrid dynamics associated with the problem. To reduce the complexity associated with reasoning about contact switching, many dexterous manipulation planning methods rely on breaking down the task by choosing hand-designed manipulation primitives \cite{Omata1996,Cruciani2018DMG,Li2020LearningHC}. Often, primitives are designed for specific tasks or objects, thus requiring significant implementation effort to generalize to different scenarios.

Finger gaiting is another common method for reducing the complexity of dexterous manipulation planning by allowing only a single finger to switch contacts at any given time. Finger gaiting has most often been used for in-hand re-grasping \cite{Sundaralingam2018GeometricIR,Fan2017RealTimeFG}. Although finger gaiting can be considered a primitive, it is more general than higher-level manipulation primitives such as sliding, pushing, or pivoting.

In contrast to breaking down a task by imposing structure on motion, data-driven approaches have also been used to plan dexterous manipulation tasks \cite{nagabandi2019deep,openai}, and while the results have been impressive, they require large amounts of training data and time to generalize to different scenarios.

Of these three strategies, our proposed approach is most similar to finger gaiting, and does not rely on choosing hand-designed primitives. Unlike \cite{Sundaralingam2018GeometricIR}, we consider the task of moving an object to a goal pose and do not assume a given order in which fingers move. \cite{Fan2017RealTimeFG} assumes that the object is smooth to enable fingers to slide along its surface.

\subsection{Contact planning for locomotion and manipulation}

Finally, we address the contact planning works most directly related to this work.

First, we discuss a recent work that combines search-based planning and dynamics optimization for humanoid contact planning \cite{LinBerenson2018}, as well as an extension to that work that considers the robustness of the method to disturbances \cite{Lin2020RobustHC}. The core structure of the methods in these works is similar to TrajectoTree, first constructing and searching through a graph to find a contact sequence, and then using a dynamics optimization step to generate full trajectories given this contact sequence. However, their graph search finds contact sequences with low dynamics costs, which traditionally would involve solving a slow dynamics optimization problem to obtain the cost of every contact transition, whereas our discrete planner only cares about finding feasible solutions. Consequently, the goal of their work is to speed up the graph search by training a network to predict the optimal objective value of the dynamics optimization for a contact transition. In contrast, this work focuses on speeding up trajectory optimization by using a search-based planner.


Next, we discuss several hybrid feedback control works that reason about contact mode scheduling. These works employ a variety of strategies, including mixed-integer quadratic programming (MIQP) \cite{Marcucci2017ApproximateHM, Hogan2016FeedbackCO}, sequencing high-level primitives \cite{Woodruff2017PlanningAC}, enumerating all possible contact sequences \cite{Doshi2019}, and learning contact schedule selection with Bayesian optimization \cite{Seyde2019LocomotionPT}. Of these, only \cite{Doshi2019} considers multi-contact dexterous manipulation tasks. The authors present a hybrid differential dynamic programming (DDP) algorithm for closed-loop execution of planar pushing and pivoting primitives with frictional contact switches. Their method uses input-constrained DDP to explore and rank all feasible contact mode sequences. Unlike CITO methods, DDP requires a fixed contact mode sequence. Consequently, their approach requires enumerating all possible contact sequences, limiting the number of contact switches they can consider while still obtaining reasonable planning times. The maximum number of contact switches they consider is two. 

Finally, we highlight two recent papers that both reason about how to plan contact switches for planar dexterous manipulation tasks. The first, \cite{AceitunoCabezas2020AGQ}, reformulates the problem with a quasi-dynamic relaxation and constructs the planning problem as mixed-integer program. The second, \cite{cheng2020contact}, uses rapidly-exploring random tree guided by contact modes. Both methods plan a diverse set of dexterous manipulation tasks, but neither consider full finger kinematics, and instead assume that the contact points will always be reachable.

\section{Method}
First, we use a discrete contact sequence planner that only considers grasp stability and kinematics to find a sequence of contact switches for the fingers given a desired object trajectory. Next, we formulate the contact transition trajectory optimization by combining the high-level plan with CITO by constraining the contact locations to those given by the plan, thereby allowing only one finger to break contact with the object at any given time. Additionally, we use the object poses from the high-level plan as incremental goal poses in the cost function. Fig. \ref{fig:method} shows an overview of our method.

TrajectoTree makes several assumptions. The discrete 
planner allows only one finger to make and break contact with the object at any given time. It also relies on the user to define a maximum number of allowed contact switches. We discuss the implications of this in the conclusion. In the contact transition trajectory optimization, we use the same number of time steps for each contact switch, but this can be modified by adjusting the constraints. The duration for each contact switch can even be included as decision variables within the optimization.

\begin{table}[h]
\centering
\caption{Primary notation}
\begin{tabular}{l|l}
\textbf{Notation} & \textbf{Definition}  \\ \hline
$n_f$             & Number of fingers in hand   \\ \hline
$n_q$             & Number of joints in hand   \\ \hline
$n_d$             & Number of degrees of freedom of object \\ \hline
$l$  & \makecell[l]{Number of contact force components transmitted \\through $n_f$ contacts} \\ \hline
$x \in \mathbb{R}^{n_v}$  & Object position and orientation \\ \hline
$q \in \mathbb{R}^{n_q}$  & Joint angles \\ \hline
$\tau \in \mathbb{R}^{n_q}$  & Joint torques \\ \hline
$\lambda \in \mathbb{R}^{l}$  & Contact forces, expressed in local contact frames \\ \hline
$G(x,q) \in \mathbb{R}^{n_d \times l}$  & Grasp matrix \\ \hline
$J(x,q) \in \mathbb{R}^{l \times n_q}$  & Hand Jacobian \\ \hline
$p_f \in \mathbb{R}^{2}$           & Position of fingertip $f$, expressed in object frame \\ \hline
$\bs{\gamma}$  & Vector of slack variables \\ \hline
superscript $^*$                & Denotes a reference quantity from high-level plan
\end{tabular}
\label{table:notation}
\end{table}

\subsection{Contact sequence planner}

\begin{figure}[t]
    \centering
    \includegraphics[width=0.9\linewidth]{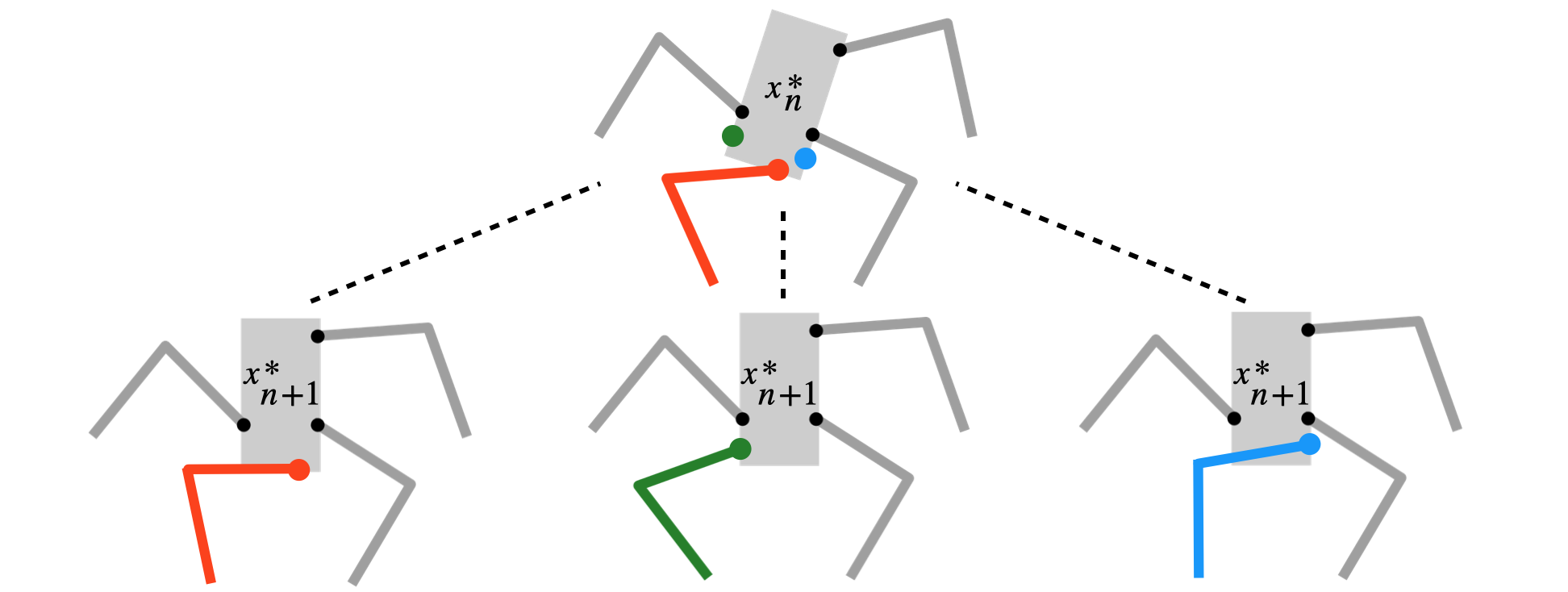}
    \caption{An example illustrating the tree-search. To get neighbors of the top node, increment the object pose, determine a free finger, shown in red, and choose new contact points based on a fixed set of contact displacements. Here, the free finger can either stay at the red contact point, or move to the green or blue contact points. Staying at the red point or moving to the blue point would result in collisions with the object. Thus, the only valid node to expand next is that with the free finger at the green point.}
    \label{fig:graph}
\end{figure}

To find a feasible finger contact sequence for a given object trajectory, we use depth-first search with a heuristic to construct and search through a tree. The planner takes as input an $N$-length object pose trajectory $(x^*_0, ..., x^*_{N-1})$ and initial contact point locations for each finger, $p^*_0$. A node $z$ in the tree consists of an object pose, finger joint configurations, and finger contact point locations on the object. To construct a tree, we expand nodes that are kinematically feasible and have a stable grasp. To expand a node, we find its neighbors by exploring a discrete set of contact switches for each finger not needed to maintain grasp stability. We outline our search algorithm with pseudo-code in Algorithm \ref{alg:search}, and Fig. \ref{fig:graph} illustrates each of its components with a simple example. Below, we describe each of the functions in Algorithm \ref{alg:search}.



\begin{algorithm}
\newcommand\mycommfont[1]{\footnotesize\ttfamily\textcolor{gray}{#1}}
\SetCommentSty{mycommfont}
\DontPrintSemicolon
\SetAlgoLined
\SetKwInOut{Input}{input}
\SetKwInOut{Output}{output}
\Input{$(x^*_0, ..., x^*_{N-1})$, $p_0^*$}
\Output{$\mathcal{T} = (V,E)$, Planner success status}
 $V_{\text{open}} \leftarrow \{z_{\text{start}}\}$,
 $V_{\text{closed}} \leftarrow \emptyset$ \\ 
$\mathcal{T} \leftarrow (V \{z_{\text{start}}\}, E \{\text{none}\})$ \tcp{Initialize tree}
\While{$V_{\text{open}} \neq \emptyset$}{
\tcp{Get deepest node from open set}
  $z_{\text{current}} \leftarrow \textsc{SelectNode}(V_{\text{open}})$ \\
  $V_{\text{open}} \leftarrow V_{\text{open}} \setminus z_{\text{current}}$ \tcp{Remove from open set} 
  $V_{\text{closed}} \leftarrow V_{\text{closed}} \cup z_{\text{current}}$ \tcp{Add to closed set} 
  \If{$\textsc{IsFeasible}(z_{\text{current}})$}{
    \If{$x^*_{\text{current}}$ equals $x^*_{N}$}{
    \tcp{Exit if object goal pose reached}
        \Return{$\mathcal{T}$, True}
    }
    \tcp{Add the neighbors to tree}
    $Z_{\text{near}} \leftarrow \textsc{GetNeighbors}(z_{\text{current}})$ \\ 
    \For{$z \in Z_{\text{near}}$}{
        $\mathcal{T} \leftarrow \textsc{Insert}(z, z_{\text{current}})$\\
        $V_{\text{open}} \leftarrow V_{\text{open}} \cup z$
    }
  }
 }
 \Return{$\mathcal{T}$, False}
 \caption{Contact sequence planner}
 \label{alg:search}
\end{algorithm}

\textsc{SelectNode}($V_{\text{open}}$): 
From the nodes in the open set $V_{\text{open}}$ deepest in the tree, return the node with the lowest heuristic value. As a heuristic for choosing feasible nodes, we use the deviation of the second finger joint from its nominal joint angle of 45 degrees, guided by the intuition that the more extended fingers are, the more likely a contact configuration will be infeasible. While we found that this heuristic helped improve search times marginally, it can be swapped with other heuristics or removed entirely.

\textsc{IsFeasible}($z$): Determine if node $z$ is valid, and should be expanded, or a dead-end. It is a dead end if the object is not in frictional form closure \cite{SpringerGrasping}, or the contact points are not reachable by the fingers. We check for frictional form closure by solving the linear program formulated in \cite{SpringerGrasping} and check for kinematic feasibility with inverse kinematics. Return \textit{False} if $z$ is a dead end and \textit{True} otherwise. 

\textsc{GetNeighbors}($z$): Return all neighbor nodes of a feasible node $z$. Given the current node's object pose, $x^*_n$, the object pose of all neighbor nodes is $x^*_{n+1}$. To get a neighbor node $z_{\text{near}}$ of $z$, first determine all free fingers in $z$. A finger is ``free'' if it is not needed to maintain frictional form closure on the object if the other fingers remain fixed on the object. Next, choose a new contact point location for that free finger by displacing its current contact location along the object's surface by some displacement chosen from a pre-defined, fixed set of displacements $D_{\text{cp}}$. The set $D_{\text{cp}}$ includes the zero displacement, meaning that the free finger will remain at its current contact location. Then, create a new neighbor for each displacement in $D_{\text{cp}}$. Only one finger can perform a contact switch between two nodes, so each neighbor node will have the same set of contact points as the parent node, $z_{\text{current}}$, except for one contact point which may have changed.

\textsc{Insert}($z_{\text{new}}$, $z_{\text{current}}$): Add $z_{\text{new}}$ to $V$, and add the edge between $z_{\text{new}}$ and its parent $z_{\text{current}}$ to $E$.

Algorithm \ref{alg:search} outputs path of $N$ nodes from the initial to final object pose, where each node contains the object pose $x^*_{n}$, joint angles $q^*_{n}$, and contact point locations in the object frame $p_{n}^*$ for \mbox{$n=0,...,N-1$}. At most one finger can perform a contact switch at any given time, meaning that with four fingers in total, any two consecutive nodes in the path must have three common contact points. We call a pair of consecutive nodes in a path a transition segment. In TrajectoTree, we enforce these transition segments in CITO by constraining contact point locations.

\subsection{General contact-implicit trajectory optimization} \label{ssec:cio}

We formulate the general CITO problem for planar dexterous manipulation. A discussion of extending this formulation to 3D can be found in \cite{PosaCIO}. In Section \ref{sec:contact-transition-to}, we augment this formulation with constraints from the contact sequence plan to arrive at the contact transition trajectory optimization used in TrajectoTree. Table \ref{table:notation} contains the primary notation.

We find trajectories for the object pose, finger joint configurations and torques, and corresponding contact forces by solving a CITO problem of the form
\begin{align}\label{eqn:opt}
&\underset{\{\mbf{x},\mbf{q},\dot{\mbf{x}},\dot{\mbf{q}},\bs{\tau},\bs{\lambda},\bs{\gamma}\}}{\text{minimize}}\;\;\;
F(\mbf{x},\bs{\tau}, \bs{\lambda}, \bs{\gamma})\\
& \quad \text{s.t.} \quad \;\;\; \text{dynamics constraints (\ref{eq:obj_dyn}), (\ref{eq:hand_dyn})}\nonumber\\[-3pt]
& \quad \quad \quad \quad \text{friction cone constraint (\ref{eq:con_fc})}\nonumber\\[-3pt]
& \quad \quad \quad \quad \text{complementarity constraints (\ref{eq:con_comp}), (\ref{eq:con_fgp1})}\nonumber\\[-3pt]
& \quad \quad \quad \quad \text{path constraints for each decision variable,}\nonumber
\end{align}
where $F(\mbf{x}, \bs{\tau}, \bs{\lambda}, \bs{\gamma})$ is a quadratic tracking objective function of the form
\begin{equation} \label{eqn:cost}
\begin{split}
F(\mbf{x},\bs{\tau},\bs{\lambda}, \bs{\gamma}) &= \sum_{k=0}^{M} (x_k - x_{\text{goal}})^T Q (x_k - x_{\text{goal}})\\
& \quad + \tau_k^T R \tau_k + \lambda_k^T L \lambda_k + ||\bs{\gamma}||_1,
\end{split}
\end{equation}

where $\mbf{x} = (x_0, \ldots, x_M)$ is the object pose trajectory, and $\mbf{q}$, $\bs{\tau}$, and $\bs{\lambda}$ are defined similarly as the joint angle trajectory, joint torque trajectory, and contact force trajectory, respectively. $L$, $Q$, and $R$ are weight matrices, and $\bs{\gamma}$ is a vector of slack variables, which we use to relax several equality constraints to improve convergence of the optimization problem \cite{Manchester2017}.

The constraints in (\ref{eqn:opt}) are as follows. 
We use trapezoidal collocation \cite{Kelly2017AnIT} to discretize the finger and object dynamics,
\begin{alignat}{2}
M_{\text{obj}}(x_k)\Ddot{x_k} &= G(x_k, q_k) \lambda_k + g_{\text{obj}}(x_k), \label{eq:obj_dyn}\\
M_{\text{hand}}(q_k)\Ddot{q_k} &= \tau_k - J^T(x_k,q_k)\lambda_k. \label{eq:hand_dyn}
\end{alignat}
In these constraints, $g_{\text{obj}}$ is the vector of gravitational forces on the object, and $M_{\text{obj}}$ and $M_{\text{hand}}$ are the object and hand mass matrices, respectively. We omit the Coriolis and centrifugal terms in (\ref{eq:hand_dyn}) because these are negligible in a quasi-static setting with low-mass manipulators. Higher-order collocation methods, like Hermite-Simpson or orthogonal collocation, can be used instead of trapezoidal collocation to improve solution accuracy, as \cite{Patel2019ContactImplicitTO}.

We model contact in the optimization problem with friction cone constraints and complementarity constraints:
\begin{alignat}{2}
\mu \lambda_{n,k} - |\lambda_{t,k}| & \geq 0 \label{eq:con_fc}\\
\gamma_i - \phi(q_{k})\lambda_{n,k} &\geq 0 \label{eq:con_comp}\\
-\gamma \leq (J\dot{q}_k - G^T \dot{x}_k)_{\{x,y\}} \lambda_{n,k} &\leq \gamma \label{eq:con_fgp1}\\
\lambda_{n,k} & \geq 0\\
\phi(q_k) & \geq 0 \label{eq:con_phi}\\
\bs{\gamma} &\geq \bs{0} \label{eq:con_slack_var}.
\end{alignat}
This ensures that contact forces are zero when the fingertips are not in contact with the object. Equations (\ref{eq:con_fc}) -- (\ref{eq:con_slack_var}) apply for each finger $f = 1,...,n_f$ separately. To minimize notation clutter, we drop the additional subscript $f$ and use $q$ and $\lambda$ to denote joint angles and contact forces for a single finger $f$.
The vector of contact forces $\lambda$ is comprised of normal and tangential components $\lambda_n$ and $\lambda_t$, respectively. Equation (\ref{eq:con_fc}) constrains the contact forces to lie within the planar friction cones of an object with coefficient of friction $\mu$. The function $\phi(q)$ can be thought of as a signed-distance field of the object, which enforces a non-penetration constraint with (\ref{eq:con_phi}) and only equals zero when the fingertip is in contact with the object. Equation (\ref{eq:con_comp}) is the complementarity constraint on normal contact forces, relaxed with a slack variable. We include another relaxed complementarity constraint, (\ref{eq:con_fgp1}), to constrain each fingertip to remain fixed at their contact points when on the object; these equations constrain the $\{x,y\}$ components of a fingertip's velocity, in the world frame, to be equal to the $\{x,y\}$ components of the contact point velocity, also in the world frame.  We solve the problem in (\ref{eqn:opt}) with the constraints in  (\ref{eq:obj_dyn}) -- (\ref{eq:con_slack_var}) using IPOPT \cite{ipopt}.

\subsection{Constraining CITO to the high-level contact sequence} \label{sec:contact-transition-to}

Finally, we formulate the contact transition trajectory optimization by combining the contact sequence plan with CITO. Given a high-level contact sequence plan, we modify the cost function and introduce additional task-specific constraints that specify which fingers are free and fixed, as well as the contact point locations of each finger. For a high-level plan with $N-1$ transition segments, we fix each segment to be $\hat{M}$ time-steps long, making the full trajectory $M = (N-1)\hat{M}$ time-steps long. Each of the $s=0,...,(N-2)$ transition segments consists of two consecutive nodes.

First, we introduce a cost function that uses object poses $(x^*_1, ..., x^*_{N-1})$ given by the high-level plan as incremental goal object poses in the running state cost, resulting in a cost function of the form
\begin{equation} \label{eqn:cost2}
\begin{split}
\hat{F}(\mbf{x},\bs{\tau},\bs{\lambda}, \bs{\gamma}) = \sum_{j=0}^{M} \tau_j^T R \tau_j + \lambda_j^T L \lambda_j + ||\bs{\gamma}||_1 + \\
\sum_{n=1}^{N-1} \sum_{k=0}^{\hat{M}}  (x_{(n-1)\hat{M} + k} - x_{n}^*)^T Q (x_{(n-1)\hat{M} + k} - x_{n}^*).
\end{split}
\end{equation}

If segment $s$ has a free finger $f'$, we constrain the contact points to be fixed at locations specified by the high-level plan at the first and last time-step of the segment, and keep the complementarity constraints for the other time-steps with constraints of the form

\vspace*{-\baselineskip}
\begin{alignat}{2}
\text{FK}(q_{s\hat{M},f'}) &= p_{s,f'}^* \label{eq:con_free_cp1}\\
\text{FK}(q_{(s+1)\hat{M}-1,f'}) &= p_{(s+1),f'}^*\\
\text{(\ref{eq:con_comp}), (\ref{eq:con_fgp1})} \text{ with } k &\in \{s\hat{M}+1,...,(s+1)\hat{M}-2\}.
\end{alignat}

The fixed fingers $f$ are constrained to be fixed at locations specified by the high-level plan for the entire segment $s$ with equality constraints of the form

\vspace*{-\baselineskip}
\begin{alignat}{2}
\text{FK}(q_{k,f}) &= p_{s,f}^* \label{eq:con_fixed_cp} \quad \text{for} \quad k &\in \{s\hat{M},...,(s+1)\hat{M}-1\}
\end{alignat}
where $\text{FK}(q_{k,f})$ is the forward kinematics that computes the fingertip position of finger $f$ at time-step $k$. In our implementation, we relax the contact point equality constraints with slack variables, but omit them here for brevity.

\begin{figure}[t]
    \centering
    \includegraphics[width=\linewidth]{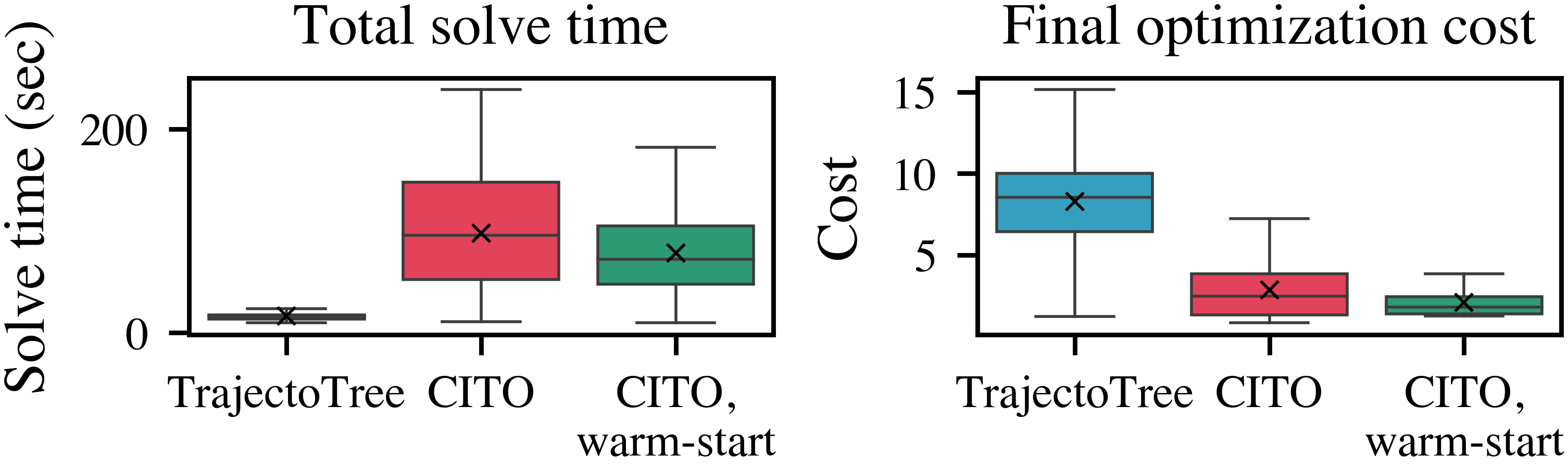}
    \caption{\label{fig:time_and_cost}Total planning times and final optimization cost of TrajectoTree and baselines across 60 different goal rotations randomly sampled between -$\pi$ and $\pi$ radians. We present the data with standard boxplots and denote means with crosses. TrajectoTree reaches a solution faster (left), but typically with a higher cost (right) due to the additional contact constraints.}
\end{figure}
\begin{figure*}[t]
    \centering
    \includegraphics[width=\linewidth]{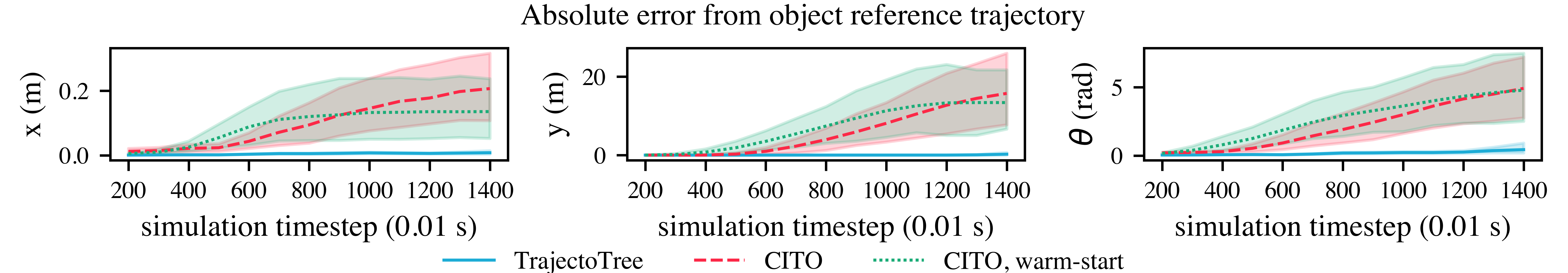}
    \caption{Mean absolute error from object reference trajectory for $x$, $y$, and $\theta$ when executing trajectories with a controller in simulation for 60 sampled goals, with 95\% confidence interval bands. TrajectoTree plans trajectories that are more robust to execute in a physics simulator, as compared to pure CITO methods. The baseline methods frequently drop the object, shown by large $y$ and $\theta$ errors in the middle and right plots, while TrajectoTree does not.}
    \label{fig:tracking}
\end{figure*}

Solving (\ref{eqn:opt}) with (\ref{eqn:cost2}) as the cost function and (\ref{eq:con_free_cp1}) - (\ref{eq:con_fixed_cp}) in place of (\ref{eq:con_comp}) and (\ref{eq:con_fgp1}) results in trajectories that adhere to the contact sequence in the high-level plan. With this formulation, we solve one optimization problem for the entire $M = (N-1)\hat{M}$ length trajectory. Additionally, we include the boundary constraints 
\begin{alignat}{2}
x_0 &= x_0^* \label{eq:con_x0}\\
q_0 &= q_0^* \label{eq:con_q0}\\
\dot{x}_{s\hat{M}} , \dot{x}_{(s+1)\hat{M}-1} & = 0 \;\;\forall\;\; s \in \{0,...,N-2\} \label{eq:con_dx}\\
\dot{q}_{s\hat{M}} , \dot{q}_{(s+1)\hat{M}-1} & = 0 \;\;\forall\;\; s \in \{0,...,N-2\},
\label{eq:con_dq}
\end{alignat}
where (\ref{eq:con_x0}) and (\ref{eq:con_q0}) constrain the initial object pose and joint angles of the entire trajectory to those specified by the first node in the high-level plan. Equations (\ref{eq:con_dx}) and (\ref{eq:con_dq}) constrain the object and joints to be stationary at the first and last time-step of each segment. We do not constrain the final object pose or joint angles of each segment to those specified in the high-level plan, thereby allowing for some deviation.

\subsection{Controller} \label{sec:controller}
To execute trajectories planned by TrajectoTree in simulation, we track the fingertip trajectories in Cartesian space using the following simplified impedance controller \cite{wuthrich2020trifinger} with additional gravity compensation for the fingers (time index omitted for brevity)
\begin{equation} \label{eq:ctr2}
\tau = J^T\Big(k_p(p_{\text{ref}}-p)+k_v(\dot{p}_{\text{ref}}-\dot{p})+\lambda_{\text{ref}}\Big) + g_{\text{hand}},
\end{equation}
where $p_{\text{ref}}$, $\dot{p}_{\text{ref}}$, and $\lambda_{\text{ref}}$ are the reference fingertip positions,  velocities, and contact forces from the trajectory optimization solution, $g_{\text{hand}}$ is the gravity compensation vector, and $k_p$ and $k_v$ are hand-tuned controller gains.

\section{Experiments \& Results}
We show that TrajectoTree achieves faster planning times than other CITO baselines. Additionally, we show that when executing the planned trajectories in a PyBullet simulation environment, we are able to more closely track object trajectories planned by TrajectoTree.

\subsection{Experimental Setup}
In our experiments, we consider contact sequences of length $N=10$, trajectories of length $M=10.8 \text{ seconds}$, and 0.1 second optimization time steps. For these results, we fix the initial pose of the object and randomly sample 60 object goal orientations between -$\pi$ and $\pi$ radians. In our experiments, we obtain the object pose trajectory for the contact sequence planner by linearly interpolating between initial and final object pose. We use the same initial contact configuration for all trials. The object we consider is a 20cm$\times$10cm rectangular object with coefficient of friction $\mu = 0.7$ and mass \mbox{$m = 50$ grams}.

\subsection{Baselines}
We compare TrajectoTree to these CITO baselines:

1) ``CITO'': General CITO formulation from Section \ref{ssec:cio}, initializing decision variables such that the system is in static equilibrium throughout the entire trajectory. We hypothesize that this formulation will produce highly dynamic trajectories prone to dropping the object during execution.

2) ``CITO, warm-start'': General formulation from Section \ref{ssec:cio}, initializing object pose and joint angles to those given by the high-level plan. We hypothesize that compared to the first baseline, this initialization should move solutions closer to a non-dynamic manipulation sequence.

For all three methods, we use cost function (\ref{eqn:cost2}) with the same values for $L$, $Q$, and $R$.

\subsection{Planning Speed and Solution Quality}
We show that TrajectoTree achieves faster total planning times than the baseline methods. For methods that use the contact sequence planner, we consider total planning time to be the sum of the search time and the trajectory optimization solve time. Fig. \ref{fig:time_and_cost} (left) reports the total planning times across 60 trials for each method. TrajectoTree achieves an average planning time of 14 seconds. ``CITO'' and ``CITO, warm-start'' achieve average planning times of 98 seconds and 76 seconds, respectively. The planning times of TrajectoTree have a significantly narrower interquartile range compared to those of the baselines, suggesting that our method also performs more consistently across various goal poses. 

This improved planning speed comes at the expense of finding trajectories that have a higher cost in comparison with the baselines. This is expected, as TrajectoTree solves an optimization problem that is the same as for the baselines, but with additional constraints corresponding to the desired contact sequence. Specifically, TrajectoTree constrains particular fingers to remain in contact with the object via typical equality constraints, rather than allowing them to make and break contact via complementarity constraints. Because TrajectoTree is more constrained than the baselines, the optimal objective value is guaranteed to be no lower than the baselines, and will be higher than the baselines if these additional constraints are active, as evidenced in Fig.~\ref{fig:time_and_cost} (right).  What is crucial to note, however, is that the trajectories found by TrajectoTree are considerably more robust to dropping the object when executed in a physics simulator, as shown in Fig.~\ref{fig:tracking} and detailed in sec~\ref{sec:controller}.  


A comparison of the two baseline methods also demonstrates that the choice of the initial guess impacts solutions, as other works \cite{PosaCIO, Dai2014} have also found; however, our results suggest that imposing additional constraints, as done in TrajectoTree, can significantly influence the solution cost and planning time. While is it clear that these additional constraints will increase the trajectory cost, it may be surprising that these extra constraints can also significantly reduce solution time and lead to more robust plans in practice.  

\subsection{Executing trajectories in simulation}
We show in simulation that controllers which track trajectories found with TrajectoTree are significantly less likely to drop the object than when tracking trajectories planned with the baseline methods. We execute trajectories in a planar PyBullet environment by planning a trajectory once and tracking the reference fingertip positions with the low-level controller described in Section \ref{sec:controller}. We use the same hand-tuned controller gains for all methods. 
We compare executing trajectories planned by all three methods, across the same randomly chosen goal poses, and show the mean absolute tracking errors for object position and angle in Fig. \ref{fig:tracking}.

Although we only perform closed-loop tracking of fingertip positions, the controller is able to maintain relatively small tracking errors on object pose when tracking trajectories planned by TrajectoTree. In contrast, tracking the trajectories planned by the baseline methods results in much larger errors and frequently dropping the object, as shown by the dramatically increasing errors in the $y$ dimension (gravity points in the negative $y$ direction in our simulation). This reinforces that imposing the additional constraints given by the high-level plan, which only considers kinematic feasibility and grasp stability, moves the CITO towards solutions that are of higher cost, but also more robust during execution.

\section{Conclusion}
This work demonstrates the utility of augmenting CITO methods with discrete planning. For dexterous manipulation tasks, we show that CITO methods are most suitable when augmented with a discrete contact sequence planner that reasons about kinematic constraints and grasp stability. Using a contact sequence planner enables us to impose task-specific constraints in the optimization, which not only dramatically reduces the planning time, but also results in trajectories that, when executed in a physics simulator with a low-level tracking controller, produce more robust object manipulation.

We only demonstrate TrajectoTree in a planar scenario and recognize that applying it to a 3D task may be difficult, due to the challenges that come with dealing with 3D orientations in trajectory optimization and the poor scaling of the contact sequence planner.
Although we fix the parameters of the contact sequence planner in our experiments, we discuss how varying these parameters would affect our method. We fix the length of the object trajectories to $N=10$ which corresponds to tree depth, and fix the number of contact point offsets to 17 which corresponds to tree branching factor. The length of the object trajectory equals the maximum number of contact switches allowed in a high-level plan, and with $N=10$ maximum switches, we find that our planner is able to find feasible solutions for object goal angles between -$\pi$ and $\pi$ radians. Planning goals farther from the initial object pose would require increasing $N$, resulting in longer search times.
One alternative to formulating the contact sequence planning problem as a tree search is to pose it as a mixed-integer program (MIP); however, our discrete planner would likely outperform most generic MIP solvers, not only because it is built to only find feasible, as opposed to optimal, solutions, but also because it admits the use of heuristics to speed up the search. 
While we acknowledge the discrete planner presented in this work scales poorly with goal distance, this is tangential to the main point of this work, which is the utility of augmenting CITO with discrete planning. Finding scalable methods for planning discrete contact sequences remains an interesting direction for future work.

\bibliographystyle{unsrt}
\bibliography{citations}

\end{document}